# Creating Artificial Students that Never Existed: Leveraging Large Language Models and CTGANs for Synthetic Data Generation

MOHAMMAD KHALIL, Centre for the Science of Learning & Technology (SLATE), University of Bergen, Norway

FARHAD VADIEE, Centre for the Science of Learning & Technology (SLATE), University of Bergen, Norway

RONAS SHAKYA, Centre for the Science of Learning & Technology (SLATE), University of Bergen, Norway

QINYI LIU, Centre for the Science of Learning & Technology (SLATE), University of Bergen, Norway

In this study, we explore the growing potential of AI and deep learning technologies, particularly Generative Adversarial Networks (GANs) and Large Language Models (LLMs), for generating synthetic tabular data. Access to quality students' data is critical for advancing learning analytics, but privacy concerns and stricter data protection regulations worldwide limit their availability and usage. Synthetic data offers a promising alternative. We investigate whether synthetic data can be leveraged to create artificial students for serving learning analytics models. Using the popular GAN model- CTGAN and three LLMs- GPT2, DistilGPT2, and DialoGPT, we generate synthetic tabular student data. Our results demonstrate the strong potential of these methods to produce high-quality synthetic datasets that resemble real students' data. To validate our findings, we apply a comprehensive set of utility evaluation metrics to assess the statistical and predictive performance of the synthetic data and compare the different generator models used, specially the performance of LLMs. Our study aims to provide the learning analytics community with valuable insights into the use of synthetic data, laying the groundwork for expanding the field's methodological toolbox with new innovative approaches for learning analytics data generation.

## 1 Introduction

Student learning data is a core component of three related areas of study and practice: data-driven decision-making, educational data mining, and learning analytics [42]. Learning analytics (LA) relies on data to achieve its goals, which often involve supporting students through various interventions, such as making predictions based on machine learning models and providing personalised learning. However, the field of LA has been engaged in ongoing debates about the ethical and practical challenges of collecting student data to further develop LA tools and models [48].

Authors' Contact Information: Mohammad Khalil, Centre for the Science of Learning & Technology (SLATE), University of Bergen, Bergen, Norway, mohammad.khalil@uib.no; Farhad Vadiee, Centre for the Science of Learning & Technology (SLATE), University of Bergen, Bergen, Norway, farhad.vadiee@uib.no; Ronas Shakya, Centre for the Science of Learning & Technology (SLATE), University of Bergen, Bergen, Norway, ronas.shakya@uib.no; Qinyi Liu, Centre for the Science of Learning & Technology (SLATE), University of Bergen, Bergen, Norway, qinyi.liu@uib.no.

In the context of LA, data collection may include information that students prefer not to disclose [52]. Although students generally trust their institutions to manage their data in a responsible way, some institutions engage in unnecessary data collection practices to train LA machine learning models [30, 43], and storing the data for extended periods of time [50]. Additionally, practitioners in the field face challenges in obtaining suitable datasets for scaling LA systems and testing their generalisability due to privacy concerns and data protection regulations, an issue that is increasingly attracting attention from the community [13, 16].

Another issue that the field faces, particularly with empirical studies, is data quality and data scarcity [28]. LA can be highly impactful; investing in data quality measures and data sharing mechanisms are key to ensure that these efforts produce reliable and actionable insights [32]. Achieving a high level of utility (e.g., high predictions accuracy) requires comprehensive data to generate insights and interventions that support student learning on a large scale [28, 31, 56]. Issues such as incomplete datasets, data inaccuracies, and biases in data collection can undermine the validity of developing and evaluating LA models. Insufficient training and validation data for student-supporting machine learning models in LA may result in flawed predictions, misleading conclusions, and underprivileged data-driven decision-making [25]. According to [55], the insufficient data problem is particularly severe in LA, with 71% of predictive model studies being conducted using small sample sizes ($N < 50$). This limits the potential of LA to optimise and support student outcomes. Therefore, addressing data scarcity issues is essential to advancing the field and ensuring that LA remains functional and robust.

Synthetic data generation, an advanced methodology that mimics real data to generate artificial data with similar structure and statistical properties, is a relatively novel approach. For LA, synthetic data generation could offer potential solutions to several longstanding challenges in the field. Still, synthetic data generations have not yet garnered the requisite attention from researchers and practitioners in LA and the broader educational data science community [18, 56]. The recent advancements in Artificial Intelligence (AI) modelling and deep learning, like Generative Adversarial Networks (GANs) and the exceptional capabilities of Large Language Models (LLMs), may enable synthetic data generation to offer a significant opportunity to address data limitations (e.g., data scarcity, scalability) encountered in many of the LA applications.

In this paper, we explore creating artificial students via leveraging LLMs and GANs to generate artificial data that replicate the characteristics and structure of real educational data of students, particularly tabular data [1]. Tabular data remains one of the most prevalent and foundational data formats in LA [31]. Our research evaluates the utility of this artificial data to real student data (i.e., in terms of resemblance, general utility, and machine learning performance of synthetic and real tabular data) to support LA modelling across various scenarios. With utility in mind, our paper aims to bridge a key gap in LA and the broader educational data science field by allowing researchers to explore and test hypotheses across different scenarios, while minimising the ethical and logistical constraints associated with using actual student data. More specifically, we explore the following research questions (RQs):

**RQ1**: To what extent can synthetic student data be leveraged and evaluated to real data and predict actual educational outcomes?

**RQ2**: How do Large Language Models (LLMs) compare to GAN-based models in terms of generating synthetic data for educational data science?

The contributions of our work are as follows: (1) We provide a comprehensive evaluation of the synthetic student data to assess its overall utility to real data for the field of LA.; (2) We introduce a new measure, the Synthetic Data Integrity

---

[1]Tabular data is organised in tables where rows represent records or observations and columns correspond to attributes or variables. A common format for such data is comma-separated values (CSV) files.

Score (*SDIS*) as a comprehensive metric to assess synthetic data performance. (3) We demonstrate that LLMs can be a viable alternative to GAN-based synthetic tabular data generation methods, and in some cases, LLMs outperform these methods. And (4) Our study highlights that synthetic data can have high-quality standards and be used in the LA field as a scalable and practical alternative to real data. This inspires potential solutions to scalability issues, data quality, and data scarcity for LA and the general educational data sciences research.

## 2  Background

### 2.1  Synthetic Data in LA

Synthetic data generation has been communicated early in education through student simulation [51]. For LA, synthetic data generation has gained attention from the community for not a long time as a method for data augmentation and privacy protection, though empirical work remains scarce. One of the earliest studies on synthetic data in LA was by [5], who argued that it could play a key role in enriching student data. However, it is only in the past five years that the practical application of synthetic data in LA has begun to grow [18]. For instance, [56] conducted a comparative evaluation of different synthetic data generation models, including statistical approaches like Gaussian Copula and deep learning models such as CTGANs. Using a large dataset from three Australian courses, they assessed the performance of the used models based on regression accuracy and Root-Mean-Square Error (RMSE). The findings were promising which suggested that synthetic data generation can enhance the supervised machine learning performance of LA models.

Another example by [38], who generated synthetic data using a variety of synthetic data methods and evaluated the results by considering two active learning strategies on three unbalanced datasets. Moles and colleagues [38] showed that synthetic data can address the minority class prediction inaccuracies imposed by limited student data. In [22] observed that a significant challenge in LA fairness research is the scarcity of publicly available, unfair datasets due to privacy concerns. To address this limitation, they employed synthetic data generation techniques to create missing datasets from scratch. Meanwhile, [31] applied synthetic data generation methods to mitigate privacy concerns for LA datasets. In [31] results showed that publishing synthetic data can protect the privacy of real data while addressing the issue of data scarcity in the LA field. The aforementioned literature demonstrates that synthetic data provides a promising method for replicating real data in the field of LA.

However, only high-quality synthetic data can, partially, replace real data. From the literature review, we found that previous studies have been relatively scattered in their evaluation of generating high-quality synthetic data for LA, often focusing on various factors such as privacy; an example of such studies is [31]. Consequently, the exploration of a key area, utility, remains insufficient. This paper aims to address this gap by providing a more comprehensive evaluation of synthetic data for this purpose. Our paper also aims to take a significant step forward in advancing the scalability of LA. By generating high-quality synthetic LA datasets that closely resemble real datasets in terms of their predictive performance, this research aims to address the data scarcity faced by LA practitioners. While previous studies in the LA field have primarily relied on statistical distribution methods and deep learning methods for synthetic data generation, recent advancements in LLMs have demonstrated their potential for generating synthetic tabular data [8]. However, to date, there has been no comprehensive evaluation of LLMs for this specific task within the LA domain.

## 2.2 Synthetic Data Generation

Synthetic data is defined as "data that has been generated using a purpose-built mathematical model or algorithm, with the aim of solving a (set of) data science task(s)" ([23], p. 5). The main types of synthetic data include tabular, time series, text, images, video, or audio simulation. All descriptions of synthetic data generation in this paper are tabular data.

There are two primary approaches to generating synthetic data: statistical distribution and deep learning methods [23]. Statistical distribution methods, such as Bayesian networks, rely on mathematical models to represent probability distribution of data. Deep learning methods, particularly GANs, use neural networks to learn the underlying patterns of the data and generate new samples. GANs have shown promising results in the field of LA. For example, in [38] experiment, the GAN-based method outperformed other methods like SMOTE, achieving nearly a 50% improvement in prediction accuracy on some datasets. Within deep learning methods, Conditional Tabular Generative Adversarial Networks (CTGANs) have emerged as a highly effective method for generating structural tabular data, particularly in scenarios where large-scale data collection is impractical or privacy concerns are paramount [53]. CTGAN provides a scalable method for generating high-quality synthetic datasets, which can significantly improve the performance of machine learning models across LA [6, 31, 56]. Therefore, in this paper, we select CTGAN as a representative GAN-based method for further experiments. It will be used as one of two approaches to generate synthetic data.

Our second approach to generating synthetic data involves LLMs. LLMs have recently gained substantial attention in both academia and industry, with ChatGPT and Google Gemini being widely recognized examples. Additionally, LLMs have significantly enhanced the performance of various natural language processing (NLP) tasks, opening up new possibilities for automating tasks that humans traditionally did. Bubeck et al [10] predict that the impressive performance of LLMs could potentially lead to the development of Artificial General Intelligence (AGI) in the future.

In LA, LLMs have been used, for example, to predict challenging moments from students' discourse [49], provide real-time feedback [41] and as an intelligent tutoring systems to support students in higher education [26].

In general, LLMs have been extensively explored for generating synthetic images and text, often outperforming other methods [24]. However, their use for generating synthetic tabular data has been far less common. Recently, though, studies have started to investigate the potential of LLMs in tabular data generation [9, 54]. One notable success is the recently developed framework GReaT (Generation of Realistic Tabular data) by [8], whose experimental results show that LLM performance on various datasets is comparable to, if not better than, leading GAN-based methods such as CTGANs. Breugel & Schaar [9] posit that LLMs could significantly impact the generation of synthetic tabular data, potentially transfiguring how tabular data is utilised in science and machine learning. Breugel & Schaar [9] argue that LLMs can be better equipped to reason about the underlying distribution of variables and generalise to new relationships within tabular datasets. Still, exploring tabular data generation using LLMs remains limited. Breugel & Schaar [9] concluded that further exploration of LLMs and benchmarking specifically designed for synthetic tabular is needed.

Our review of the literature reveals no comprehensive study in LA or educational data science comparing LLM-based models with GAN-based methods for synthetic data generation. This paper addresses this gap.

In our study to use LLMs for synthetic data generation for tabular data, we adopt Borisov et al's GReaT [8]. GReaT utilises autoregressive generative LLMs to sample synthetic and high utility and resemblance tabular data. We selected three relatively lightweight and open-source LLMs to engine the creation of synthetic data, namely GPT2, DistilGPT2, and DialoGPT. Below we provide a brief history of each.

- **GPT2**- is a pre-trained transformer model developed by OpenAI in 2019, designed using a causal language modelling objective [46]. GPT2 is trained on a large corpus of English text in a self-supervised manner which enables it to generate coherent and contextually relevant language outputs. We used the smallest version of GPT2, which has 124 million parameters, and it can be accessed from the Hugging Face website[2].

- **DistilGPT2**- a distilled version of GPT2 with 82 million parameters, was developed by Hugging Face [47]. Like GPT2, DistilGPT2 can be used for text generation. This model was developed through knowledge distillation, a process designed to produce a faster, lighter version of GPT2. Knowledge distillation is described in [47].

- **DialoGPT**- is a state-of-the-art, large-scale pre-trained model designed for generating responses in multi-turn conversations. Developed by Microsoft, the model was trained on 147 million multi-turn dialogues sourced from Reddit discussion threads. The details of the model can be found in [57].

### 2.3 Synthetic Data Evaluation

Evaluating synthetic data is essential to determine its usefulness. Since this paper focuses on utility, it is important to select appropriate metrics to assess the quality of synthetic data. The simplest evaluation approach involves using descriptive statistics, such as comparing the mean, median, and standard deviation with those of real data. While similar values may suggest that the synthetic data resembles the real data, this method can be misleading [17]. As shown by [4], datasets with nearly identical descriptive statistics can have very different distributions, meaning descriptive statistics alone may not provide an accurate assessment.

Another promising approach to evaluating synthetic data involves measuring its resemblance to real data through mathematical metrics. These metrics have been applied in some LA research (e.g., [31]) and more widely across other fields of synthetic data generation [23, 58]. Among the most frequently used methods are *Wasserstein Distance* (*WD*), *Jensen-Shannon Divergence* (*JSD*), and the *Chi-squared* test, which are detailed further in Section 3.4. Each of these metrics provides a different lens through which the alignment of synthetic and real data can be understood.

More recently, researchers such as [2] and [45] have introduced innovative measures—such as *Quality*, *Detection*, and *Utility* to evaluate the utility of synthetic data. These methods go beyond basic similarity (i.e., resemblance), but also offer deeper insights into the replicability and efficiency of synthetic data to real data.

This is particularly important for assessing the utility of synthetic data in machine learning applications. By examining the performance of models trained on synthetic data in comparison to those trained on real data, we can gauge whether the synthetic data adequately captures the underlying structure of the real data. If performance is comparable, it suggests that the synthetic data is not only statistically similar but also functionally useful. This approach has gained attention as one of the most robust ways to evaluate the generalisation potential of synthetic data[23].

## 3 Methodology

To address the two research questions in our study, we develop a methodological pipeline as depicted in Figure 1. This pipeline comprises several key components: datasets description (Section 3.1), environment and setup (Section 3.2), synthetic data generation methods (Section 3.3), and finally, the evaluation phase (Section 3.4).

### 3.1 Datasets Description

We selected five tabular datasets from four diverse sources based on LA systems, each varying in size and scale to enhance representativeness. This selection was further refined by incorporating datasets with varying degrees of

---

[2] https://huggingface.co/openai-community/gpt2

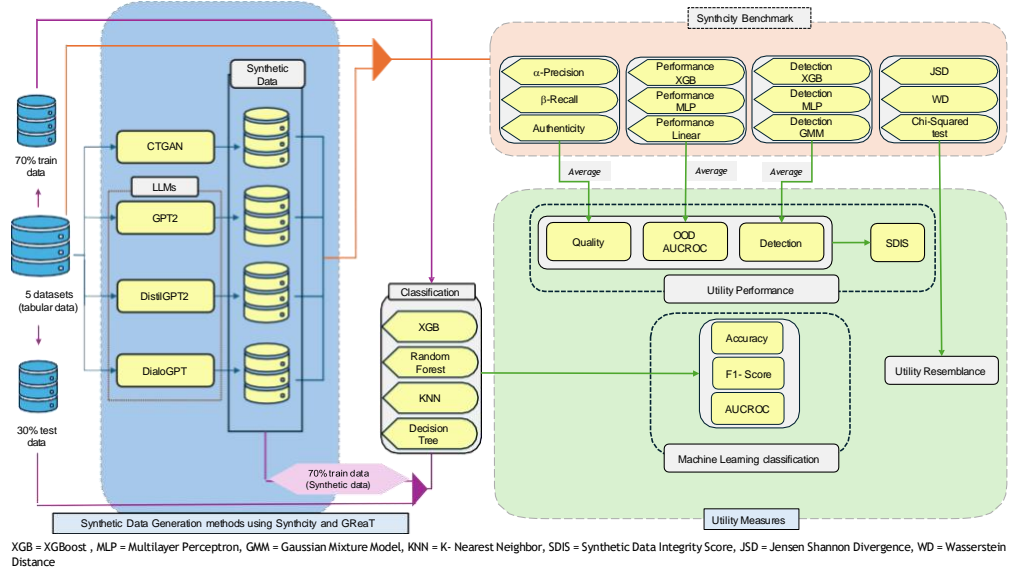

XGB = XGBoost , MLP = Multilayer Perceptron, GMM = Gaussian Mixture Model, KNN = K- Nearest Neighbor, SDIS = Synthetic Data Integrity Score, JSD = Jensen Shannon Divergence, WD = Wasserstein Distance

Fig. 1. The pipeline of the methods and measures used for generating and evaluating synthetic data in LA balance, as recommended by [19] for synthetic data generation. Notably, two of the datasets (i.e., B1 and B2) are identical in structure but differ in size, allowing us to examine the impact of dataset scale on synthetic data generation.

Prior to utilising our target datasets for synthetic data generation and subsequent evaluation -where they are partitioned into 70% for training and 30% for testing as shown in Figure 1- a series of preprocessing steps were performed. These include random sampling, computation of inner products, imputation of missing values (NaNs), normalisation of features, encoding of categorical variables, addition of new columns, and feature engineering to optimise the datasets. Brief descriptive information about each dataset after being preprocessed, including its class and source, is provided in Table 1. The following describe the datasets in details:

Table 1. Brief description of the selected LA datasets after preprocessing. Labels *Cont.* for Continuous variables and *Categ.* for Categorical variables. Target variable is used later for machine learning prediction.

| ID | Datasets | Year | #Attributes | #Records | Target Variable | Cont. | Categ. |
|----|----------|------|-------------|----------|-----------------|-------|--------|
| A | Student performance dataset from UCI (Mathematics) | 2014 | 33 | 395 | G3 (Final grade) | 16 | 17 |
| B1 | 10% of Open university dataset "studentinfo" | 2015 | 12 | 3259 | final_result | 3 | 9 |
| B2 | 30% of Open university dataset "studentinfo" | 2015 | 12 | 9778 | final_result | 3 | 9 |
| C | Subset of MOOC dropout dataset | 2023 | 27 | 2000 | Truth (Dropout) | 27 | 0 |
| D | Student Performance and Engagement Prediction eLearning datasets | 2020 | 16 | 486 | Final Exam | 15 | 1 |

- Student performance dataset from UCI (referred as Dataset A, [11]). This dataset contains 649 records and 33 columns, capturing information on student achievement from two Portuguese schools. Dataset A is divided into two subsets, focusing on the subjects of Mathematics and Portuguese language. For our experiment, we used only the Mathematics subset. The dataset includes demographic features and school grades.

- Open University Learning Analytics dataset (OULAD studentinfo, [27]). This dataset contains 32,593 rows and 12 columns, consisting of anonymized student data with unique identifiers, demographic information, and results from seven different courses in a Learning Management System (LMS). The columns include the number of attempts students made in specific modules, their credit scores, demographic details, education level, and final results. Due to the dataset's large size, which can be resource-intensive for computation, we created two subsets for our experiment: the first (referred to as Dataset B1) consisting of a random 10% selection (i.e., 3,259 records) and the second (referred to as Dataset B2) comprising a random selection of 30% of the dataset (i.e., 9,778 records).
- MOOC dropout dataset (referred as Dataset C, [15]). The dataset contains 12,944,862 rows and 8 columns, representing web tracking logs from 247 unique MOOC courses on the Chinese XuetangX platform. For our experiments, We transformed the MOOC dataset to aggregate action types into a single-row structure per student to simplify classification training. Additionally, we added a column capturing total activity time as an engagement indicator. This preprocessing step was performed to improve the accuracy of predictive machine-learning models for the dropout variable. We randomly selected a subset of 2,000 records.
- Student Performance and Engagement Prediction eLearning datasets (referred to as Dataset D, [39]) merge two sources: clickstream data and performance. The data was extracted from the Owl LMS at a North American university. Performance data includes students' grades for assignments, quizzes, and exams. Both datasets were cleaned and merged using student IDs, resulting in a unified dataset containing activity logs and grades.

## 3.2 Environment and Setup

The experiments in this study for the synthetic data generation and evaluation were conducted on Google Colab using two premium accounts to access the A100 GPU. The A100 GPU, part of NVIDIA's Ampere architecture, provided the necessary computational power for generating and evaluating synthetic data. The training of LLMs coupled with the evaluation of synthetic data to assess its structural accuracy and predictive performance requires significant computational resources. The A100 GPU serves invaluable in enabling us to carry out this work.

We use the Synthcity library [44] mainly to generate synthetic data and benchmarks. Synthcity is a Python library designed to generate and evaluate synthetic tabular data and can automatically calculate about 25 metrics.

## 3.3 Synthetic Data Generation Methods

We generate synthetic data using four generative models: CTGAN, DistilGPT2, DialoGPT, and GPT2. We integrate these models into our experiment setups using the Synthcity library. The generative models learn the statistical patterns from the real data by processing the original dataset as input. For CTGAN, this means directly working with the real dataset's tabular structure, learning the distribution of both categorical and numerical features, as well as the relationships between columns. On the other hand, the GReaT framework transforms each row of the tabular dataset into a text-based representation and feeds it into DistilGPT2, DialoGPT, and GPT2 in order to generate similar data [8].

We generate synthetic data by using both the LLMs and CTGAN of the same size (i.e., number of rows and columns) as the real datasets to ensure consistency for downstream tasks. Each column value is synthetically produced based on the patterns learned from the real data which preserve key relationships between the dataset's features while mimicking realistic variability. This implies that augmenting the datasets is feasible; however, for our research work, we aimed to create a consistent number of dataset attributes and records. Our intended approach maintains the dataset's structure, which we believe is essential for evaluating and training predictive models as close as to the real data.

*3.3.1 Parameter tuning.* For CTGAN, we use the default plugin values provided by the Synthcity library. The three most important parameters are 2,000 training iterations, 200 Batch sizes, and 0.001 learning rate. These parameters influence the model's ability to learn the data distribution. We maintained these default values to provide a standard implementation of CTGAN for our experiments.

To generate synthetic data using the three LLMs, we incorporate the GReaT framework [8] as a plugin into the Synthcity library. The GReaT framework leverages advanced pre-trained transformer language models to produce high-quality synthetic tabular data. GReaT converts each row of the dataset into text and uses it to fine-tune the target LLM. However, the default implementation of GReaT imposes a limit on text length, which poses challenges for datasets with a large number of features that require longer text representations, as is the case in our study. To overcome this limitation, we modified the GReaT framework by implementing our own constructor function. This allows us to fine-tune the LLMs with longer texts. We use the Hugging Face library within GReaT to download and fine-tune the three LLMs.

We adjusted the training parameters for each LLM and dataset as the following: For the MOOC dataset (Dataset C), we set the number of epochs (i.e., iterations of the training data) to 30 and the batch size (number of samples) to 32. For the Student Math dataset (Dataset A), we used 100 epochs with a batch size of 32. For the Student Info 10% dataset (Dataset B1), we set the number of epochs to 50 and the batch size to 32, while for the Student Info 30% dataset (Dataset B2), we used 20 epochs with the same batch size. Finally, for the Student Performance and Engagement dataset (Dataset D), we adjusted the value of epochs to 30 and the batch size to 32. These adjustments allowed us to manage computational resources to meet our requirements while ensuring sufficient training for each model.

## 3.4 Evaluation of Synthetic Data Generation

In this study we use a variety of evaluation metrics to provide a solid utility examination of synthetic data following the guidelines of [2, 23, 31, 45, 56]. In detail, we use *WD*, *JSD*, *Chi-squared*, *Quality*, *Detection*, and *Utility* metrics tests for evaluating the utility of performance and resemblance for synthetic data. To avoid confusion between the *Utility* metric proposed by [2] and [45] and the broader concept of utility as a general evaluation of synthetic data for machine learning models, we rename the *Utility* metric from [2] and [45] to *OOD AUCROC* [3].

To provide an overview of *Quality*, *Detection*, and *OOD AUCROC*, we introduce the *Synthetic Data Integrity Score* (*SDIS*). For further assessment of the performance of synthetic data in machine learning, we generated synthetic datasets of the same size as the real datasets. Classifiers were trained on 70% of the data—either real or synthetic—and tested on 30% of the real data, ensuring consistent training data volumes for a fair comparison. The evaluation metrics include *Accuracy*, *AUCROC* score, and *F1-score*. Below is a brief description of each metric:

- **Wasserstein Distance** (*WD*), also known as the Kantorovich-Rubinstein metric, measures the distance between probability distributions in a given metric space [40]. *WD* is used for addressing continuous data to maintain resemblance in distribution shape [58]. A smaller value of *WD* indicates similar distributions between real and synthetic data, while a larger *WD* indicates a larger difference between the two.

- **Jensen Shannon Divergence** (*JSD*), is a metric used to measure the resemblance between two probability distributions [36]. *JSD* values range between $0 - 1$, where a larger value indicates greater dissimilarity. *JSD* has been used to measure the difference in the probability mass distribution between real and synthetic datasets and offers stable results for categorical data.

---

[3]Out-of-Distribution Area Under the Receiver Operating Characteristics Curve.

- **Chi-squared Test** is a statistical method used to determine whether there is a significant difference between the expected and observed frequencies in categorical data. *Chi-squared* test is represented in terms of *p*-value. When the *p*-value score is 0, the distributions are different, when the value is 1, the distributions are identical [12]. *Chi-squared* test is used for categorical features.

- **Quality**, intuitively, is a measure used to measure the quality of synthetic data relative to real data. It is calculated as the average of three metrics: *α-precision*, *β-recall*, and *Authenticity* [33, 45]. In [2], *α-precision*, *β-recall*, and *Authenticity* are used to evaluate the realism, coverage, and generalizability of generated synthetic data. *α-precision* measures the similarity between synthetic data and real data, a high value of *α-precision* shows how closely the synthetic data corresponds to the real data. *β-recall* measures the diversity within the generated synthetic data. The generated data must be diverse enough to get the existing variability in the real data. *Authenticity* measures how well the model generated synthetic data which are not just the copies of the real data. *Authenticity* helps determine the generalizability of the model. The *Quality* value ranges from 0 to 1, with 1 representing the optimum and 0 the worst.

- **Detection** measures how easy it is to distinguish synthetic data from real data using machine learning models. This metric is calculated as the average of *AUCROC* scores from three post-hoc classifiers: XGBoost, Multi-Layer Perceptron, and Gaussian Mixture Models. *Detection* ranges between 0 and 1, where 0 means the synthetic data is indistinguishable [33, 45].

- **OOD AUCROC** is a metric that evaluates the generalisation performance of classifiers trained on synthetic data when tested on real-world data from a different distribution. Specifically, it measures the ability of a model trained on synthetic data to distinguish between classes in out-of-distribution (OOD [4]) real data, providing insights into the utility of the synthetic data for practical machine learning applications. We report the performance of average *OOD AUCROC* for three classifiers: XGBoost, Multi-Layer Perceptron, and Linear, as suggested in [33]. Note that higher *OOD AUCROC* is better for the generated synthetic data.

- **The Synthetic Data Integrity Score (SDIS)** is a comprehensive metric that we propose to provide a complete overview of synthetic data performance. *SDIS* encapsulates the overall effectiveness of the data by integrating three key aspects—*Quality*, *Detection*, and *OOD AUCROC*—into a single score. The *SDIS* value ranges from 0 to 1, with 1 representing the best performance and 0 the worst performance.

$$SDIS = \frac{Quality + OOD\ AUCROC + (1 - Detection)}{3}$$

- **Classifiers**: We employed widely used machine learning predictive classifiers, including XGBoost, Random Forest, K-Nearest Neighbors (KNN), and Decision Trees to train on synthetic data and evaluate their performance on real data. The classifier measures used for this purpose are: 1) *Accuracy* to measure the ratio of correct predictions to the total number of predictions made by a model. 2) *F1-score* is the weighted mean of precision and recall. And 3) *AUCROC* is a performance metric for machine learning models.

## 4 Findings

This section presents the results of the generation of the synthetic tabular data as well as the detailed evaluation of the five LA tabular datasets. The results form grounding to answer our research questions (*RQ*1 and *RQ*2). Due to limited space, all the generated synthetic data by the four generative models CTGAN, GPT2, DistilGPT2, and DialoGPT as well as the evaluation results can be accessed via the GitHub repository [5].

---

[4]Out-of-distribution data refers to data drawn from a distribution that differs from the training data distribution, often encompassing new or unseen scenarios that the model has not encountered before.

Before going into the details of our findings, we provide Fig. 2, a t-distributed stochastic neighbour embedding (t-SNE) plot for Dataset B2, as an example, which depicts the comparison between synthetic data generated by each generative model and the real data. t-SNE is a statistical method for visualising high-dimensional data, enabling clearer observation of similarities between synthetic and real data distributions [20, 34]. The results from t-SNE show strong capability to resemble real data in terms of overall distribution. Due to space limitations, t-SNE plots for other datasets are available in the provided repository.

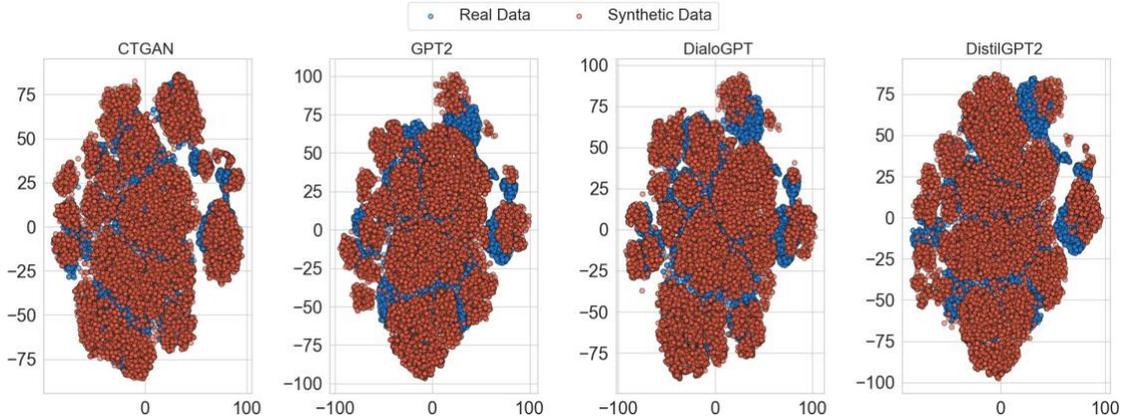

Fig. 2. t-SNE plots showing the similarity between synthetic data generated by each generative model and the real data for Dataset B2. Best viewed in colour.

Following the study's pipeline in Fig. 1, we present the findings of the evaluation for 1) utility performance using *Quality*, *Detection*, *OOD AUCROC*, and *SDIS* in Section 4.1; 2) utility resemblance using *WD*, *JSD*, and *Chi-squared* test in Section 4.2, and 3) the evaluation results of the machine learning classification using *AUCROC*, *F1-score*, and *Accuracy* of the synthetic data in Section 4.3. The evaluation of both utility performance and utility resemblance metrics were performed twice. We present the mean value of each metric.

## 4.1 Evaluation of Utility Performance

Table 2 compares the four models of CTGAN, DistilGPT2, GPT2, and DialoGPT across the different datasets (A, B1, B2, C, and D) using three metrics: *Quality*, *Detection*, and *OOD AUCROC*.

CTGAN demonstrates higher *Quality* scores for the datasets B1, B2, and C which the averages are well above 0.64. In contrast, GPT2 and DialoGPT show strong performance, with GPT2 attaining the highest *Quality* scores of 0.6259 in Dataset D and DialoGPT with a score of 0.6443 in Dataset A. On the other hand, DistilGPT2 produces lower *Quality* scores compared to other models, which implies that the reduction in model size may compromise its ability to generate data that mimics real distributions. In [33], it is suggested that scores in this range indicate high-quality synthetic data, positioning our result within a favorable spectrum.

Table 2. Average values of *Quality*, *Detection* and *OOD AUCROC* across LA datasets and various generative models. Bold indicates best performance. Underline indicates second best performance. (↑) indicates for the given metrics higher the score higher the performance, (↓) indicates for the given metrics lower the score beNer the performance. ± value represents standard error of the mean

| Datasets | Metrics | CTGAN | DistilGPT2 | GPT2 | DialoGPT |
|---|---|---|---|---|---|
| A | *Quality* (↑) | <u>0.6403 ± 0.1387</u> | 0.5951 ± 0.125 | 0.616 ± 0.1474 | **0.6443 ± 0.1223** |
| | *Detection* (↓) | **0.6454 ± 0.0709** | 0.795 ± 0.1531 | <u>0.6503 ± 0.0927</u> | 0.673 ± 0.0761 |
| | *OOD AUCROC* (↑) | 0.5335 ± 0.0853 | <u>0.5975 ± 0.0539</u> | 0.4069 ± 0.1429 | **0.5983 ± 0.0605** |
| B1 | *Quality* (↑) | **0.6405 ± 0.1598** | 0.6018 ± 0.2077 | <u>0.6182 ± 0.1121</u> | 0.5938 ± 0.2131 |
| | *Detection* (↓) | <u>0.6232 ± 0.1514</u> | 0.657 ± 0.1717 | **0.5744 ± 0.0749** | 0.662 ± 0.169 |
| | *OOD AUCROC* (↑) | 0.5866 ± 0.0541 | 0.5854 ± 0.0298 | <u>0.5904 ± 0.0223</u> | **0.6002 ± 0.0288** |
| B2 | *Quality* (↑) | **0.656 ± 0.1622** | 0.5875 ± 0.196 | <u>0.6122 ± 0.1007</u> | 0.6087 ± 0.0947 |
| | *Detection* (↓) | 0.6211 ± 0.1071 | 0.6829 ± 0.1592 | <u>0.5963 ± 0.085</u> | **0.5815 ± 0.0807** |
| | *OOD AUCROC* (↑) | **0.6051 ± 0.039** | 0.5776 ± 0.0513 | <u>0.5925 ± 0.0346</u> | 0.5739 ± 0.0506 |
| C | *Quality* (↑) | **0.6451 ± 0.1523** | 0.6363 ± 0.1064 | <u>0.639 ± 0.107</u> | 0.6294 ± 0.127 |
| | *Detection* (↓) | 0.6826 ± 0.1248 | 0.6785 ± 0.1227 | <u>0.6504 ± 0.1236</u> | **0.6328 ± 0.1116** |
| | *OOD AUCROC* (↑) | **0.6012 ± 0.1012** | 0.4615 ± 0.1663 | <u>0.4718 ± 0.1747</u> | 0.4563 ± 0.1621 |
| D | *Quality* (↑) | <u>0.5918 ± 0.1623</u> | 0.561 ± 0.1153 | **0.6259 ± 0.1078** | 0.5707 ± 0.1456 |
| | *Detection* (↓) | 0.6321 ± 0.1433 | 0.5854 ± 0.0737 | <u>0.5703 ± 0.092</u> | **0.5413 ± 0.0636** |
| | *OOD AUCROC* (↑) | -0.0868 ± 0.1818 | -0.7052 ± 0.5997 | <u>-0.0682 ± 0.2196</u> | **-0.0088 ± 0.1116** |

DialoGPT performance in terms of *Detection* score is better compared to the other generative models, with its best score in Dataset D with 0.5413, where lower *Detection* score means better performance. The synthetic data generated using DialoGPT is more difficult to distinguish from real data than both CTGAN and the other LLMs. Conversely, DistilGPT2 performs the worst in this metric, especially in Dataset A with the score 0.7950, meaning it is 79.5% likely to be detected as artificial. CTGAN's performance is variable, performing well in Dataset A with a score of 0.6454 and then it is outperformed by other models in other datasets.

For *OOD AUCROC* score, DialoGPT performed best for Dataset A with the score of 0.5983, B1 with the score of 0.6002 and D with the score of −0.0088. GPT2 performs consistently well on most datasets, with its best score in Dataset B1 (0.5904). DistilGPT2 shows a varying performance. It performs well in Dataset A but exhibits a sharp drop in Dataset D. CTGAN performs well in Dataset B2 (0.6051) and Dataset C (0.6012). However, similar to other generative models it exhibits a negative *OOD AUCROC* value for Dataset D, indicating models vary in handling complex data.

Our conclusions from the three metrics (i.e., *Quality, Detection*, and *OOD AUCROC*) mapping are as the following: CTGAN generally exhibits the highest performance in terms of quality across all datasets, although its results are inconsistent in *Detection* and *OOD AUCROC*. DialoGPT achieves results comparable to CTGAN, particularly excelling in *Detection* and *OOD AUCROC*. However, its overall quality scores are generally lower than those of CTGAN. GPT-2 also shows competitive *Quality* results similar to CTGAN and maintains consistently strong performance in *Detection*, especially in datasets B1 and B2. In contrast, DistilGPT2 tends to be weaker across all three metrics compared to the other models, but it does achieve strong *Detection* scores in specific datasets, such as Dataset B2.

Note that the negative values of *OOD AUCROC* for Dataset D may give the impression that synthetic data does not perform well in terms of *OOD AUCROC*. However, in our experiments, we also observed negative values for real data of Dataset D. We have chosen not to mention the *OOD AUCROC* values for real datasets because for *Quality* and *Detection* metrics are not applicable in this case; these metrics for synthetic data are computed in comparison to real data.

Next, in Fig. 3, we depict the *SDIS* metric evaluation of the five tabular datasets for the generative models. The y-axis shows the value of *SDIS* metrics and the x-axis lists the five datasets. The figure presents that almost all the generative models perform consistently across most of the datasets maintaining *SDIS* value close to 0.5, however, the performance drops for all the models for Dataset D. Specifically, for datasets A, B1, B2, and C, CTGAN performs well with consistent high *SDIS* scores and more stable performance when compared with the other three LLMs. With respect to GPT2 and DialoGPT, both LLMs show similar performance levels, with some fluctuations across the datasets and less performance stability than the CTGAN. DistilGPT2 exhibits the lowest overall performance, particularly for datasets B2 and D, likely due to its smaller model size and fewer parameters compared to GPT2 and DialoGPT.

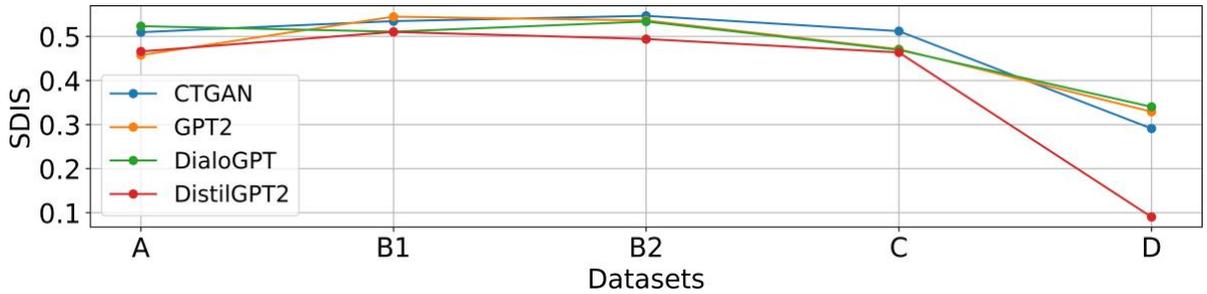

Fig. 3. *SDIS* score for each generative model and each dataset. Best viewed in colour

## 4.2 Evaluation of Utility Resemblance

On reporting to what extent synthetic data resemble the real data in distribution, we evaluated the statistical distribution of *WD*, *JSD*, and *Chi-squared* test of the synthetic data with reference to the real data (see Fig. 4). The *WD* figure reveals several interesting observations. Dataset A consistently demonstrates the highest mean *WD* scores across all generative models which suggest a relatively large divergence from the real data. This could return to the large number of continuous and categorical dimensions of the dataset which makes the distance more sensitive. While most models exhibit similar *WD* scores, datasets B1 and B2 display notably the lowest distances which as a result indicate a closer structural similarity between the synthetic and real data. Dataset D also shows relatively low mean *WD* values, except for CTGAN. Our summary of the *WD* analysis suggests that LLMs can generate synthetic data that is comparable to CTGAN in terms of resemblance.

When evaluating the mean *JSD*, the analysis shows results that differ slightly from the *WD* analysis. The mean *JSD* scores are consistently low, typically below 2% which means the resemblance for the synthetic data is very high. For Dataset A, all generative models exhibited similar performance, with *JSD* values around 0.0170. However, for datasets B1, B2, and C, CTGAN consistently outperformed the other models. Notably, on Dataset D, CTGAN showed a significant divergence, with the highest *JSD* value, indicating the worst performance among the four models. In our summary, the *JSD* analysis suggests that CTGAN is the top performer in most cases (except for Dataset D), with GPT2 closely following. While both CTGAN and GPT2 excel on multiple datasets, it is noteworthy that their distribution similarity can vary depending on dataset characteristics.

On the other hand, the *Chi-squared* test assesses the similarity between real and synthetic data in terms of *p*-value. The mean *p*-values in Fig. 4 are mostly above 0.5 for four out of five datasets. However, for Dataset D, the mean *p*-value is notably low which suggests a statistical difference between real and synthetic data as explained by [12]. For Dataset A, the *p*-values for all the generative models are in the range between 0.6 and 0.7, with CTGAN achieving the lowest score among them. For Dataset B1, CTGAN and GPT2 produce high p-values, above 0.7, while DialoGPT and DistilGPT2 yield lower values. In Dataset B2, CTGAN *p*-value slightly increases, GPT2's remains the same, but DialoGPT and DistilGPT2's values increase significantly, nearly matching GPT2's—possibly due to B2's larger size. For Dataset C, CTGAN has the highest value, followed by DistilGPT2 and DialoGPT, with GPT2 showing the lowest *p*-value. In Dataset D, CTGAN again leads with the highest value, DialoGPT and DistilGPT2 follow, but the *p*-value is relatively low and close to zero compared to other datasets. We can conclude that in terms of the *Chi-squared* test, all the generative models were nearly identical across the datasets, with the exception of Dataset D. Because the *Chi-squared* test is sensitive to categorical records, this may account for the lower *p*-values observed in the four generative models for Dataset D where the categorical records were the fewest among the other datasets.

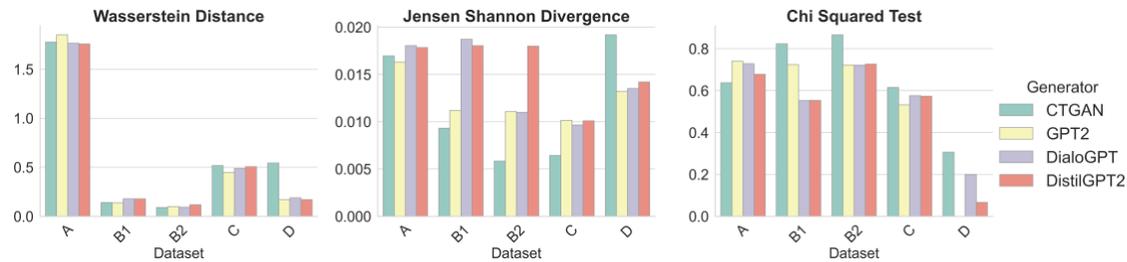

Fig. 4. The mean value for *WD* (left), *JSD* (middle), and *Chi-squared* test (right) for each dataset with reference to the real data. Best viewed in colour.

### 4.3 Evaluation of Machine Learning Classification

To check the performance of the synthetic data in machine learning experiments, we created tasks using a set of supervised learning classifiers, including Random Forest, KNN, XGBoost, and Decision Tree (see Fig. 5). Both the real and synthetic data were split into 70% for training and 30% for testing. For each generative model, we train the classifiers on 70% of the synthetic training data, utilising grid search to identify the optimal parameters. The trained models are then evaluated on the 30% of the real test data. The performance of the models were assessed by calculating *AUCROC*, *F1-score*, and *Accuracy* which enabled us to determine the quality of predictions of synthetic data in training machine learning models. These results are then compared with those obtained when the models are trained on real data. It is important to note that the classification tasks for datasets A, C, and D are binary, while the classification tasks for datasets B1 and B2 are multi-class.

Fig. 5 depicts the performance of the classifiers as plots in terms of *AUCROC* score and *F1-score* for all datasets, four classifiers, and four generative models together with real data. The circles represent the performance of real data, and various other shapes represent the synthetic data generated by different generative models. Generally speaking, the scores of the real data (represented by a circle) lie slightly within the high performance region in the plot.

For the synthetic data, in Dataset A, the synthetic data generated by GPT2, DialoGPT and DistilGPT2 are well within the highest performance region when the dataset undergoes classification task using the Random Forest. Under the XGBoost classifier, DistilGPT2 outperformed the real data, achieving an *F1-score* of around 0.92 and an *AUCROC* score nearly identical to that of the real dataset. Notably, GPT2 demonstrates the best performance among all generative models in Dataset A, even exceeding the real data when it undergoes classification tasks with the Decision Tree classifier.

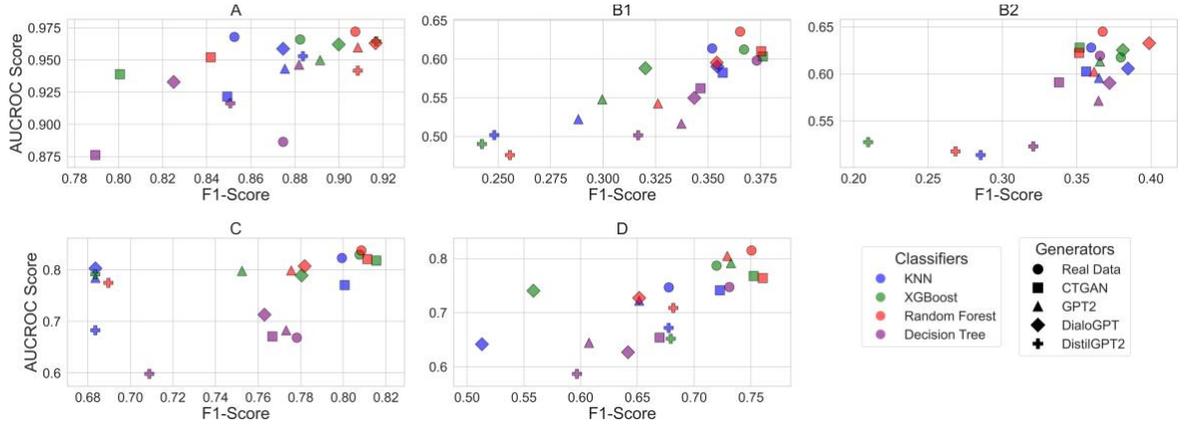

Fig. 5. *AUCROC* score and *F1-score* for each dataset, classifier, and generative model. Best viewed in colour.

In datasets B1 and B2, CTGAN and DialoGPT have comparable *AUCROC* and *F1-score*, although the overall scores in both datasets are relatively lower compared to the other datasets. DistilGPT2 shows weak performance across both datasets, with *AUCROC* score below 0.53 and *F1-score* below 0.32. GPT2 exhibits average performance, while DialoGPT and CTGAN outperform the real data in *AUCROC* during the XGBoost classification task on Dataset B2.

In Dataset C, CTGAN achieves performance as close to the real data across three of the classifiers, both in *AUCROC* and *F1-score*. However, under the Decision tree classifier, DialoGPT and GPT2 have higher *AUCROC* scores than the real data. DistilGPT2 has a low *F1-score* across all the classifiers, although the *AUCROC* score exceeds 0.75 for the Random forest and XGBoost classifiers.

In Dataset D, CTGAN has higher *F1-score* than the real data when trained using KNN, XGBoost, and Random forest classifiers. But with Decision Tree the synthetic data generated by almost all the generative models performed poorly. The synthetic data generated by DialoGPT and DistilGPT2 consistently underperforms across all the classifiers, with *AUCROC* Score below 0.75 and *F1-score* below 0.70.

The results depicted in Fig. 5 highlight the varying performance of the generative models when compared to real data in classification tasks. To provide a more comprehensive overview of the four generative models' performance, we conducted a comparative analysis of their mean *Accuracy*, *AUCROC*, and *F1-score* across the synthetic and real datasets, as illustrated in Fig. 6.

In Fig. 6, the three evaluation metrics demonstrate that the four generative models—CTGAN, GPT2, DialoGPT, and DistilGPT2—exhibit relatively strong performance compared to real data, with only marginal differences and few exceptions. While real data consistently outperforms the generative models, the differences across all metrics remain minimal which shows a robustness of the synthetic data generated by the models. Notably from the figure, CTGAN slightly exceeds the other LLMs and delivers performance comparable to real data across all three metrics for datasets B1, B2, C, and D. In certain metrics, such as *AUCROC* for datasets B1, B2, C, and D, as well as *F1-score* for datasets B1 and B2, DialoGPT and GPT2 achieved performance comparable to that of CTGAN. Among the three LLMs, DistilGPT2 consistently underperformed, with few exceptions. It is important to note that the predictive tasks for datasets B1 and B2 involve multi-label classification, resulting in lower metric values compared to the other datasets. Overall, the *Accuracy* differences between the classification machine learning models trained on synthetic data and those trained on real data are minimal—less than 7%. Surprisingly, in certain cases, the generative models trained on synthetic data outperform those trained on real data over the three metrics. This might be explained by the reduced overfitting feature enabled by the generative models, along with the balanced class distribution that offers equal representation of all data points [17, 31].

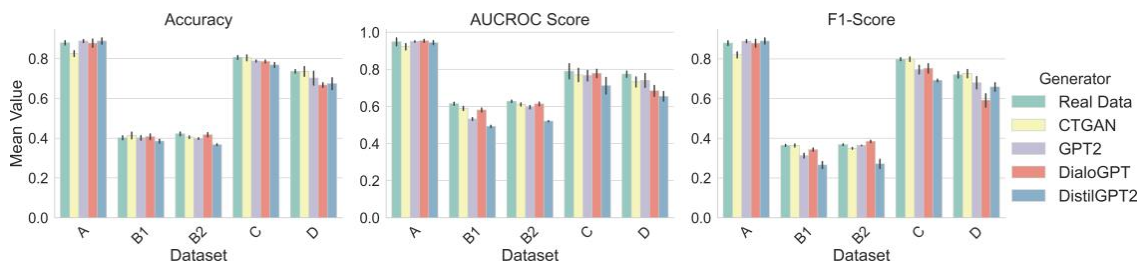

Fig. 6. The mean value for the classification machine learning evaluation metrics of *Accuracy* (left), *AUCROC* (middle), and *F1-score* (right) for all the LA datasets and the generative models. Black bar stands for the standard error of mean. Best viewed in colour.

## 5 Discussion and Conclusion

This study investigates the application and potential of synthetic data in LA by addressing two key research questions, RQ1: To what extent can synthetic student data be leveraged and evaluated to real data and predict actual educational outcomes? And RQ2: How do LLMs compare to GAN-based models in terms of generating synthetic data for educational data science? Our findings add novel insights to the current underexplored research on synthetic data for educational data science in general, and LA in particular. The study also contributes to the emerging discourse of examining the utility of synthetic data and exploring the difference between LLMs and GANs for that purpose.

Following, we address the two RQs questions of the study by designing a pipeline (Fig. 1) and conduct experiments that entail synthetic data generation and evaluation. With respect to RQ1, interestingly, the results demonstrate that the synthetic tabular datasets generated align closely with real-world datasets, particularly in terms of statistical similarity. The evaluation of utility resemblance metrics *WD*, *JSD*, and *Chi-squared* test demonstrate that the synthetic tabular data can replicate, to a large extent, the underlying statistical distribution of real educational data. The findings from CTGAN are consistent with recent studies by [56] and [31], who explored the use of GANs for generating synthetic educational tabular data and reported promising results in terms of data similarity. Another key finding that stands out from our study, which is timely and novel, is the potential of LLMs to generate synthetic tabular data. In fact this finding accords with [9] and [37]. The latter study demonstrated a comparable performance of LLMs to CTGAN in generating synthetic tabular datasets that closely resemble real-world distributions, as evidenced by our *WD* compelling scores. The minimal disparity in resemblance performance between LLMs and CTGAN models can be partially attributed to LLMs respective trained parameters size. The substantially larger number of parameters in LLMs allows them to model more intricate data patterns and dependencies; we evidenced that for DialoGPT.

The evaluation also considers three important dimensions, namely *Quality*, *Detection*, and *OOD AUCROC*, which provide a comprehensive assessment of the synthetic tabular data's realism, distinguishability and generalisability. These findings suggest that synthetic data serves as a reliable proxy for real data in certain analytical tasks, though its effectiveness may vary depending on the dataset's complexity. Additionally, we introduce the Synthetic Data Integrity Score (*SDIS*), which offers a more nuanced evaluation of synthetic data over these three dimensions.

Also with respect to RQ1, the present results show that predictive tasks of, for example, the final grade or dropout on real data generally achieves slightly better results than that of synthetic data. Yet, the differences are marginal for the deep learning generators, especially for CTGAN, and LLMs like DialoGPT and GPT2. One unanticipated result was that in some cases, synthetic data has even exceeded the performance of real data in prediction evaluation metrics such as Decision Tree, highlighting an interesting and important potential for synthetic data to delegate real data in certain predictive applications. The three plot of *Accuracy*, *F1-score*, and *AUCROC* further reinforces this finding and is supported by evidence from previous works like [14] and [21] who confirmed with fine tuning ML models for predictive tasks, synthetic data may show superior results.

Regarding RQ2, the evaluation of LLMs for synthetic tabular data generation for LA shows that LLMs can produce synthetic data comparable to CTGAN in terms of statistical properties and predictive performance. However, this finding has been met with scepticism by [54], who raised the question of whether it is theoretically possible or impossible for LLM architectures to generate certain types of synthetic datasets. As our study outcome might prove the opposite, we believe that LLMs introduce challenges related to computational efficiency and fine-tuning requirements. LLM models generally require more time and computational resources compared to traditional methods like CTGAN. Despite these challenges, the flexibility and adaptability of LLMs make them very promising engines for future synthetic tabular data generation, particularly when fine-tuned and optimised. Notably, the attention on LLMs for synthetic data generation is growing rapidly, with significant advancements such as Nvidia's recent release in August 2024 of their Nemotron LLM [3], designed specifically to enhance synthetic data generation capabilities. Nvidia's growing focus on synthetic data underscores the increasing importance of this field across various business industries.

Our study approves that determining the most effective method for synthetic data generation is complex and context-dependent [56]. Based on the performance and resemblance metrics used in this work, there is often a trade-off that varies according to specific use cases, with data type emerging as a more sensitive factor than dataset size. For example, in the OULAD dataset (B1 and B2), the larger Dataset B2, which is three times the size of B1, demonstrated better predictive accuracy, performance, and statistical similarity. This suggests that the sensitivity of the dataset is more influenced by data type, particularly categorical versus continuous, than by size alone. For those interested in integrating synthetic data generation into their research, both CTGAN and the three tested LLMs offer viable options, with marginal differences between them. However, for those focused on using LLMs in LA, DialoGPT showed the best overall performance among the models tested, followed by GPT2.

## 5.1 Study Limitations

It is important to note that this study is more focused on the quality, utility, and resemblance of the generated synthetic data, and for this reason, we exclude consideration of the privacy evaluation of the generated synthetic data. Our primary limitation is the restricted time and computational resources, which impact several areas. First, fine-tuning the LLMs is challenging which leads us to select hyperparameters like batch size and the number of epochs in a way that allows us to run experiments within our resource limits. Unfortunately, this results in suboptimal accuracy. Second, our evaluation metrics are based on two repetitions, with the mean taken as the final result. We believe that a higher number of repetitions would yield a more accurate assessment. Additionally, in this study, we exclude some popular LLMs. For instance, models like GPT3/4 are not open-source, which conflicts with our intention to use open-source models. Many LLMs, such as Llama 3 require high-end GPUs for fine-tuning, which poses another challenge when selecting LLMs. Lastly, the complexity, particularly in terms of time, increases significantly for high-dimensional and large datasets which are also limited to tabular data and excludes time-series one. As a result, we randomly select subsets of larger datasets (e.g., B1).

## 6 Future Direction

While this study provides significant insights into the use of synthetic data, several areas remain open for further research. One particularly important direction is the potential of synthetic data to improve fairness in predictive modelling (see for example [29]). Synthetic data might offer a potential to generate more balanced and equitable datasets than real datasets, which could help mitigate biases in model training. Future work is required in LA to explore methods addressing fairness issues for demographic records using synthetic data generation, as well as evaluation metrics to assess their results. Some studies have already explored this in healthtech [7] and demographic studies [1]. Additionally, future studies could focus on improving the efficiency of LLMs for synthetic tabular data generation, such as exploring different open-source LLMs and investigating methods such as zero-shot and few-shot learning to reduce computational demands. As synthetic data gains wider application in various domains, investigating new metrics beyond traditional statistical and predictive evaluations will be important to gain a more comprehensive understanding of synthetic data's impact and utility in the LA field and educational data science.

Our study's core question, which was inspired by [35] and adapted to our context, asks: Can we create artificial students with profiles so authentic they mirror reality? Can synthetic data achieve such a complex ambition? For years, LA researchers have grappled with log files, often struggling with issues of data quality, privacy, and scarcity, which hampers efforts to produce scalable, portable, and generalizable insights. This ongoing challenge leaves us to wonder if advances in AI and deep learning, particularly with technologies like GANs and LLMs, might offer a solution. We believe this innovation holds immense promise, and we approach it from an optimistic perspective, seeing synthetic data generation potential to expand the horizons of the LA field. Our study sought to travel through this possibility, though we recognise this is just the beginning.